\documentclass[10pt,twocolumn,letterpaper]{article}

\usepackage{iccv}              

\definecolor{iccvblue}{rgb}{0.21,0.49,0.74}
\usepackage[pagebackref,breaklinks,colorlinks,allcolors=iccvblue]{hyperref}
\usepackage{multirow}
\usepackage{mathtools}
\usepackage{makecell}
\usepackage{adjustbox}

\def\paperID{3674} 
\def\confName{ICCV}
\def\confYear{2025}

\title{Extending Foundational Monocular Depth Estimators to Fisheye Cameras \\ with Calibration Tokens}

\author{Rit Gangopadhyay\textsuperscript{\rm 1}\thanks{Equal contribution} , 
 Jung-Hee Kim\textsuperscript{\rm 2}\footnotemark[1] ,
 Xien Chen\textsuperscript{\rm 1}\footnotemark[1] ,
 Patrick Rim\textsuperscript{\rm 1}, 
 Hyoungseob Park\textsuperscript{\rm 1},
 Alex Wong\textsuperscript{\rm 1}\\
Yale University\textsuperscript{\rm 1}, Michigan State University\textsuperscript{\rm 2}\\
{\tt\small \{rit.gangopadhyay,xien.chen,patrick.rim,hyoungseob.park,alex.wong\}@yale.edu, kimjun84@msu.edu
}
}

\begin{document}
\maketitle

\begin{abstract}
    

We propose a method to extend foundational monocular depth estimators (FMDEs), trained on perspective images, to fisheye images. Despite being trained on tens of millions of images, FMDEs are susceptible to the covariate shift introduced by changes in camera calibration (intrinsic, distortion) parameters, leading to erroneous depth estimates. Our method aligns the distribution of latent embeddings encoding fisheye images to those of perspective images, enabling the reuse of FMDEs for fisheye cameras without retraining or finetuning. To this end, we introduce a set of Calibration Tokens as a light-weight adaptation mechanism that modulates the latent embeddings for alignment. By exploiting the already expressive latent space of FMDEs, we posit that modulating their embeddings avoids the negative impact of artifacts and loss introduced in conventional recalibration or map projection to a canonical reference frame in the image space. Our method is self-supervised and does not require fisheye images but leverages publicly available large-scale perspective image datasets. This is done by recalibrating perspective images to fisheye images, and enforcing consistency between their estimates during training.  We evaluate our approach with several FMDEs, on both indoors and outdoors, where we consistently improve over state-of-the-art methods using a single set of tokens for both. Code available at: \href{https://github.com/JungHeeKim29/calibration-token}{https://github.com/JungHeeKim29/calibration-token}; \\
\href{https://github.com/Suchisrit/CalibrationTokens}
{https://github.com/Suchisrit/CalibrationTokens}. 

 
\end{abstract}
\section{Introduction}

Three-dimensional (3D) reconstruction is a fundamental component in many spatial applications, including autonomous vehicles, extended reality (XR), robotic manipulation. Each of these applications has unique demands for the field of view (FOV), often wider than the standard (perspective) camera. To meet this need, these applications tend to be deployed on systems equipped with fisheye or other wide-angle cameras, which allows for wider coverage of the 3D environment. However, images captured by these cameras also come with substantial distortion, which arise from differences in projective geometry, where straight lines within the 3D environment or the 3D scene are preserved in perspective images but may appear curved in fisheye images. 

\begin{figure}[t]
    \centering
    \includegraphics[width=\linewidth]{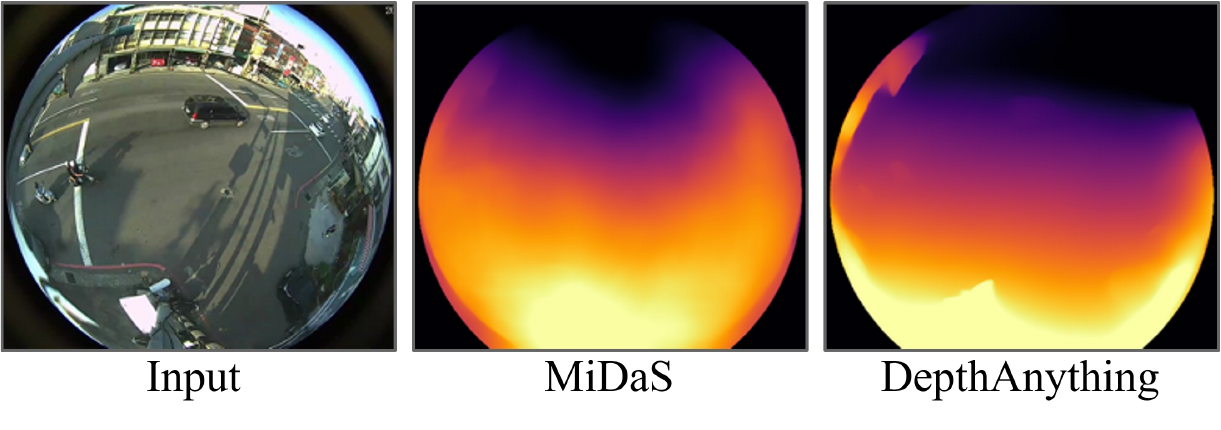}
    \vspace{-0.8cm}
    \caption{\textbf{Foundational monocular depth estimators fail on fisheye images.} Despite being trained on large-scale datasets, foundational monocular depth estimators (FMDEs) models produces erroneous outputs. The inaccurate, blurry estimates are caused by a covariate shift that stem from fisheye distortion.} 
\label{fig:bad_baseline}
\vspace{-0.4cm}
\end{figure}

Foundational monocular depth estimators (FMDEs) \cite{yang2024depthanything,piccinelli2024unidepth,ranftl2021vision} are trained on orders of tens of millions of images, enabling them to generalize across a wide range 3D scenes. However, their training data is comprised of internet images, which are predominantly captured using perspective cameras. Hence, despite being trained on large-scale datasets, FMDEs produce erroneous estimates when transferred to fisheye images (see Fig. \ref{fig:bad_baseline}). These errors stem from a covariate shift, which can be characterized by changes in camera calibration (intrinsic, distortion) parameters -- leading to differences in object appearance and their perceived depth or distance from the camera.


To address fisheye distortion, one solution is to recalibrate and undistort images or perform a map projection to some canonical reference frame. In principle, if one has the correct calibration, it is possible to re-project a fisheye image into a perspective-like view (or vice versa). In practice, however, there are several problems: (1) The calibration process itself can be error-prone and sensitive to physical perturbations in the camera system. Minor bumps, focus changes, or lens replacements can degrade or invalidate previously computed intrinsic parameters. (2) Even when re-projection is performed accurately, the transformation introduces latency and spatial artifacts (e.g., stretching, cropping, aliasing, loss). When used as a preprocessing step for existing pretrained depth estimators, these artifacts still present a covariate shift and can degrade performance. 

Another solution is to train a separate monocular depth estimator specifically for fisheye images. However, publicly available image datasets for fisheye cameras are orders of tens to hundreds of times smaller than those for perspective cameras. Hence, it is difficult to assemble sufficient data to reach the large-scale training requirement of an FMDE. Nonetheless, one can adapt or finetune existing FMDEs for fisheye imagery. While this can improve performance on fisheye images, it introduces the risk of parameter drift, where the resulting FMDEs may lose their generalizability across 3D scenes. Moreover, the resulting finetuned model becomes specialized to fisheye cameras, limiting its applicability to other camera types, which adds operational overhead in applications involving mixed camera systems, e.g., autonomous vehicles or robotics.


To address these challenges, we propose a novel approach termed \emph{Calibration Tokens}. 
Our key insight is that existing FMDEs are already capable of estimating depth for perspective images, and that errors on fisheye images are caused by a covariate shift due to differences in camera calibration and distortion. Hence, rather than retraining or finetuning the entire model, we aim to ``recalibrate'' the fisheye latent embeddings such that they become more conducive to an FMDE originally trained on perspective images. Leveraging the fact that many FMDEs \cite{piccinelli2024unidepth,yang2024depthanything,ranftl2019towards,ranftl2021vision} follow a Transformer-based architecture \cite{dosovitskiy2021an}, we will exploit the (self- and cross-) attention mechanism to modulate the latent (token) embeddings by inserting Calibration Tokens as part of the input. Therefore, the existing FMDE will remain effectively unchanged, while Calibration Tokens serve to adapt their internal representations to mitigate the covariate shift by aligning the latent embeddings of fisheye images to the distribution of latent embeddings of perspective images. This design also allows us to preserve the original image content without performing any spatial re-projection, ensuring the process is lossless in terms of the raw pixels. Our hypothesis is that by adding a small set of trainable tokens to encode the fisheye camera calibration information and utilizing them to recalibrate the latent embeddings, we will be able to reuse existing FMDEs trained on perspective images and adapt them to fisheye images without sacrificing their generalizability across diverse 3D scenes.

To train these Calibration Tokens, we propose a self-supervised objective that leverages inverse warping in the input and output spaces. FMDEs can infer high-fidelity depth maps for perspective images, so we use the perspective image depth estimates as our training target. We then induce artificial distortion on perspective images to create pairs of perspective and synthetic fisheye images with diverse fisheye distortions.  However, rather than doing the same in the output space, we undistort the fisheye depth maps to the original perspective reference frame to compute a self-supervised loss between the undistorted fisheye and perspective depth maps. By minimizing this self-supervised loss, the Calibration Tokens learn to align fisheye image embeddings to those of perspective images in the latent space, without any labels. Additionally, computing the loss in the original perspective frame allows our method to preserve the supervision signal instead of introducing artifacts.

Our approach allows us to bypass the need to compile large-scale fisheye datasets by exploiting the abundance of perspective image datasets. As our method operates in the reference frame of the input, we also avoid transformation artifacts at inference time, whether in the input or output space. Furthermore, our method preserves compatibility with perspective images: One simply needs to append or remove Calibration Tokens for FMDEs to be applied to fisheye or perspective images. We demonstrate our method on indoors and outdoors across several recent FMDEs and consistently improve over baselines. 


\begin{figure}[t]
    \centering
    \includegraphics[width=\linewidth]{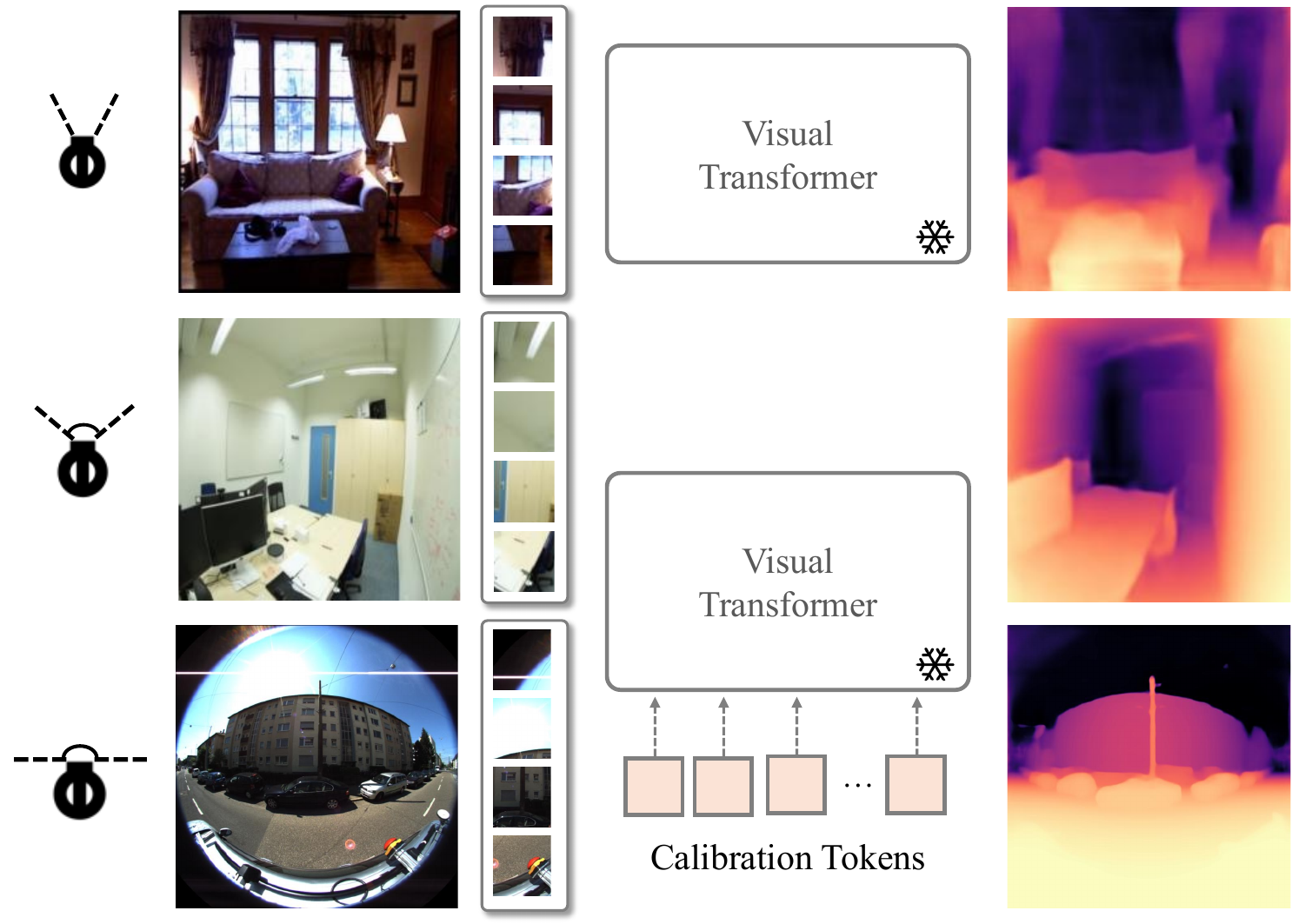}
    \vspace{-0.5cm}
    \caption{\textbf{Inference on different cameras.} Calibration Tokens enable foundational monocular depth estimators to adapt to fisheye images while maintaining performance on perspective images.} 
\label{fig:inference}
\vspace{-0.3cm}
\end{figure}

\textbf{Our contributions}: 
(1) We propose a novel approach to extend foundational monocular depth estimators (FMDEs) trained on perspective images to fisheye images. 
(2) We introduce Calibration Tokens that modulate the latent embeddings of fisheye images towards the distribution of perspective image embeddings. 
(3) We introduce a self-supervised training objective that recalibrates input perspective image to fisheye images, but ``undos'' the transformation in the output to enable loss computation on high-fidelity (perspective) depth maps inferred by FMDEs.
(4) Our approach only requires training one set of tokens to achieve state-of-the-art performance for both indoors and outdoors.


\section{Related Works}


\noindent\textbf{Monocular Depth Estimation} can be trained in a supervised or unsupervised manner. \textit{Supervised methods} \cite{eigen2014depth,eigen2015predicting,laina2016deeper,li2015depth,liu2015learning,lao2024sub,lao2024depth,yin2019enforcing,xu2017multi} minimize the difference between depth estimates and ground-truth depth maps. \cite{fu2018deep} re-formulated the problem as ordinal regression while other methods proposed architectures innovations. \cite{bhat2021adabins} partitions depth ranges into adaptive bins. \cite{chang2021transformer} incorporates an attention-based up-sample block. \cite{li2023depthformer} employs hierarchical aggregation and heterogeneous interaction modules. \cite{yuan2022neural} uses neural window fully-connected CRFs to compute energy. \cite{upadhyay2023enhancing} synthesizes perspectively accurate images to enrich training data. Additional inputs e.g., language \cite{zeng2024wordepth,zeng2024rsa,zeng2024priordiffusion}, lidar \cite{ezhov2024all,rim2025protodepth,yang2019dense,chen2024uncle}, radar \cite{rim2025radar,singh2023depth}, are used to enable metric-scale depth estimates. \textit{Unsupervised methods} \cite{choi2021adaptive,chung2025eta,guizilini20203d,liu2022monitored,park2024test,poggi2020uncertainty,tosi2019learning,watson2019self,wong2019bilateral,wong2020unsupervised,wong2021adaptive,wong2021learning,wong2020targeted,zhan2018unsupervised} minimizes photometric reconstruction error. \cite{garg2016unsupervised} frames depth estimation as a novel view synthesis problem. \cite{godard2017unsupervised} introduces a left-right consistency loss. \cite{zhou2017unsupervised} uses a pose network to enable unsupervised training on video sequences. \cite{godard2019digging} introduced auto-masking and min-reprojection loss. Additional loss terms based on visual odometry \cite{wang2018learning,fei2019geo}, iterative closest point \cite{mahjourian2018unsupervised}, surface normals \cite{yang2018lego}, trinocular assumption \cite{poggi2018learning}, and semantic segmentation \cite{guizilini2019semantically,kumar2021syndistnet} were also introduced. 
\cite{lyu2021hr} redesigned the skip connection and decoders to extract high-resolution features, \cite{zhao2022monovit} combined global and local representations and \cite{zhang2023lite} introduced a lightweight architecture with dilated convolution and attention. AugUndo \cite{wu2024augundo} leveraged invertibility of transformation groups for data augmentation.

\noindent\textbf{Foundational Monocular Depth Estimators} are trained with supervised or semi-supervised learning on large-scale datasets. MiDaS \cite{ranftl2019towards} is the first to demonstrate generalizable monocular depth estimation by compiling datasets for large-scale training. DPT \cite{ranftl2021vision} extended the approach and introduced transformers for dense predictions. Marigold \cite{ke2024repurposing} repurposes diffusion models for monocular depth estimation. DepthAnything \cite{yang2024depthanything} proposes a pseudo-labeling method to curate a large-scale dataset.
Additionally, UniDepth \cite{piccinelli2024unidepth} employs a camera self-prompting module and a pseudo-spherical output space, enabling metric-scale depth prediction across diverse 3D scenes without relying on external camera parameters. DepthPro \cite{bochkovskii2024depth} proposes a multi-scale vision transformer for metric-scale depth estimation. As all of these FMDEs are trained on perspective images, they fail to generalize to fisheye cameras. 

\noindent\textbf{Fisheye Images.} Images taken by a fisheye camera are distorted and unsuitable for use in a perspective image encoder. Existing distortion correction algorithms \cite{duane1971close, kannala2004generic} rely on different camera projection models \cite{kingslake1989history, miyamoto1964fish, stevenson1996nonparametric} to undistort images into a perspective view. However, these methods depend on camera calibration parameters, which can introduce artifacts due to calibration inaccuracies.
Recent approaches \cite{lichy2024fova,guo2025depth} demonstrate training a separate model to perform depth estimation with different camera types. They utilize an equirectangular projection to project points from different reference frames to a canonical equirectangular frame, but this can incur transformation artifacts and distortions.
Additionally, deep-learning-based methods \cite{liao2019dr, feng2023simfir} that aim to rectify distortion have been introduced. However, these methods require a large number of parameters with limited accuracy and field of view. Consequently, many recent works targeted for fisheye images involve training an entire network \cite{yogamani2024fisheyebevseg, arsenali2019rotinvmtl, zhao2024fisheyedepth} exclusively on fisheye images. Our method extends foundational monocular depth estimators to fisheye images instead.

\begin{figure*}[ht]
    \centering
    \includegraphics[width=0.99\linewidth]{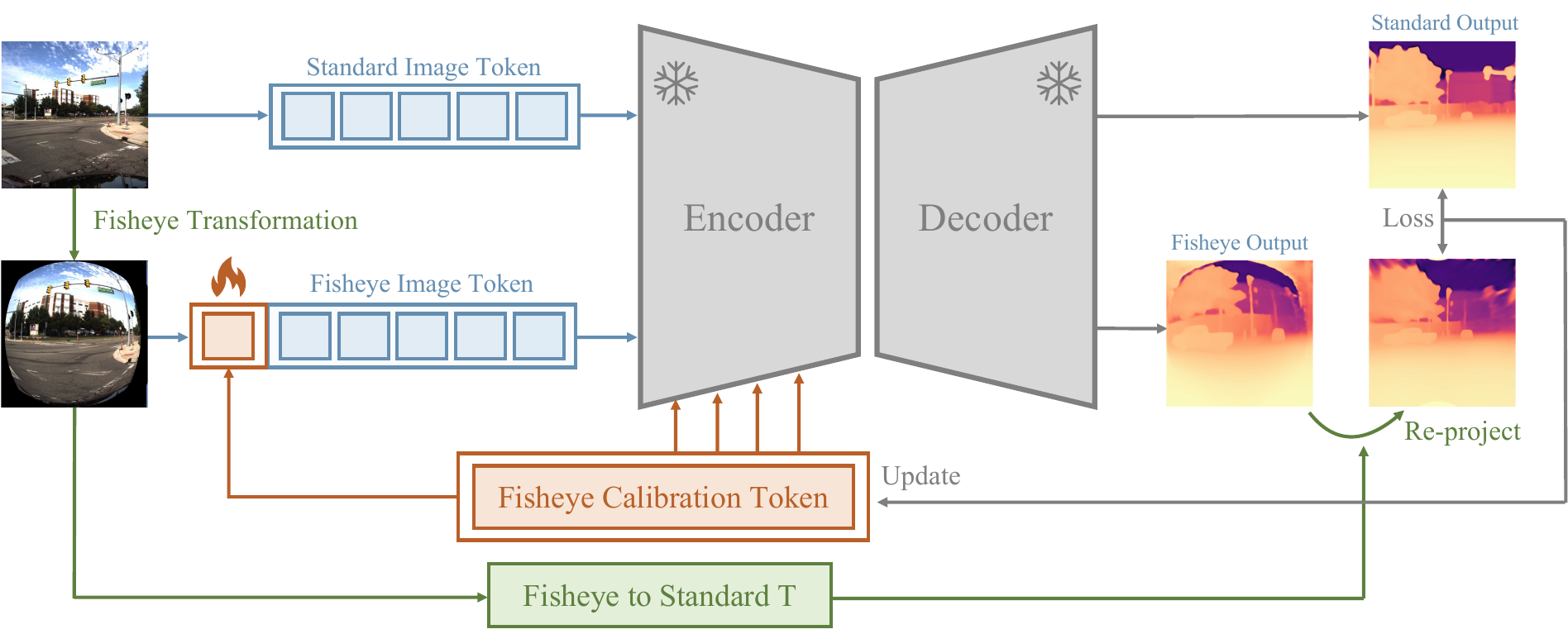}
    \caption{\textbf{Overview of our method.} We introduce a set of trainable \emph{Calibration Tokens}, which is appended to the input sequence of the fisheye image tokens. The Calibration Tokens are trained to adapt the model to produce accurate depth maps for images with various fisheye distortions. A unique fisheye calibration token is appended to the input of each new layer of the encoder.}
    \label{fig:method}
    \vspace{-0.3cm}
\end{figure*}

\noindent\textbf{Token-Based Methods.} Recent transformer-based architectures represent input images~\cite{dosovitskiy2021an} (or other modalities~\cite{xie2023sparsefusion,xia2023quadric}) as sequences of tokens. In many cases, an additional token (e.g. [CLS] token in BERT \cite{devlin2019bert} or the distillation token in DeiT \cite{touvron2021training}) is employed to aggregate information across all tokens. Such tokens can be adapted to various purposes, acting as a compact representation that ``binds'' or fuses information, e.g., \cite{yang2024binding} uses tokens to learn synthetic and real tactile response maps. Inspired by these advances, we introduce minimal trainable tokens appended to the fisheye embedding, enabling the model to ``bind'' or reconcile image distortions within a frozen backbone. Our approach is lightweight and requires no major architectural modifications, but extends foundational monocular depth estimators trained on perspective images to fisheye images.

\section{Method}

Let $I : \Omega \mapsto \mathbb{R}^3$ denote an RGB image obtained from a calibrated camera and $\Omega \subset \mathbb{R}^2$ the image space. Monocular depth estimation aims to learn a parameterized function  $h_{\omega,\psi} : \mathbb{R}^3 \rightarrow \mathbb{R}_+$ that maps an image to a depth map $d : \Omega \mapsto \mathbb{R}_+$. We assume access to a foundational monocular depth estimator (FMDE) pretrained on some large-scale dataset of perspective images. We will pair each image with (pseudo)ground truth $\tilde{d} = h_{\omega, \psi}(I)$ inferred by the FMDE to obtain a training dataset $\mathcal{D} = \{(I^{(n)}, \tilde{d}^{(n)} \}_{n=1}^{N}$.

To extend FMDEs, trained on perspective images, to fisheye images, we introduce Calibration Tokens as an adaptation mechanism. Due to the prevalence of Transformer architectures in many FMDEs, we train a set of light-weight tokens to model the change in calibration between a perspective camera and different fisheye cameras. The goal of our Calibration Token is to recalibrate or translate latent embeddings of fisheye images back to those of perspective images. Our method takes advantage of the attention mechanism inherent in FMDEs and enable Calibration Tokens to modulate the latent embeddings, thus facilitating latent alignment. The outcome is an FMDE that is capable of inferring depth for fisheye images with Calibration Tokens, and perspective images without.

\subsection{Extending FMDEs with Calibration Tokens}
Specifically, let $\phi \in \mathbb{R}^{M \times F}$ represent a set of Calibration Tokens, where $M$ denotes the number of tokens and $F$ their number of dimensions. For a given layer within the encoder $f_\omega$ of an FMDE, we will concatenate Calibration Tokens to the input sequence of patches or (embeddings) of the vision transformer: $f_{\omega}([I ; \phi])$ = $[z^{(L)} ; f_\omega(\phi)]$, where $z$ denotes the latent embeddings recalibrated by $\phi$, $L$ the last layer, and $[ ; ]$ the concatenation operation. 

As each layer denotes a separate latent space, we extend our approach to a multi-layer modulation scheme. Let $\Phi \in \mathbb{R}^{L \times M \times F}$ be the set of Calibration Tokens for each layer $l \in \{1, ..., L\}$. 
A unique set of Calibration Tokens $\phi^{(l)} \in \mathbb{R}^{M \times F}$ is appended at each encoder layer: $f_{\omega}^{(l)}[(z^{(l-1)} ; \phi^{(l)}]) = [z^{(l)} ; f_{\omega}^{(l)}(\phi^{(l)})]$ for a layer $l$. Each set of Calibration Tokens will modulate the patch embeddings for a specific layer through the attention mechanism; hence, following the convention in existing works \cite{darcet2023vision, burtsev2020memory}, we discard Calibration Tokens from the encoder output. A key insight is that the FMDE is already able to estimate high-fidelity depth maps for perspective images. We posit that the covariate shift exist in the encodings of fisheye image.  Hence, once the latent embeddings of fisheye images have been recalibrated to those of perspective images, the decoder will be able to estimate depth to similar fidelity as perspective images. Therefore, we do not utilize Calibration Tokens to modulate the decoder layers. The final estimate is obtained by $\hat{d} = g_{\psi}(z^{(L)})$, where $g_\psi$ denotes the decoder.

Since our method does not apply spatial transformations during inference, it remains entirely \emph{lossless} for input images. Additionally, it offers several efficiencies: (1) The only trainable parameters in our method are the light-weight Calibration Tokens, which consist of significantly fewer parameters than vision transformer models. Our method introduces minimal computational overhead and results in lower time and space complexity than training or finetuning a full model. (2) Our approach is backward-compatible with perspective images. By omitting our Calibration Tokens, an FMDE maintains its original depth estimation performance on perspective images. (3) At inference, camera intrinsics are not required, as the training process allows generalization across various fisheye camera intrinsics. As a result, our method eliminates the need for the arduous and error-prone calibration process after training.





\subsection{Learning Calibration Tokens}
To train Calibration Tokens, we will leverage the abundance of publicly available perspective image datasets. 
During our training, we synthesize fisheye images from perspective images by recalibrating them using artificial fisheye intrinsic and distortion parameters. This will produce pairs of perspective and fisheye images from which we can leverage self-supervision, and allows us to use a much larger training dataset than exclusively training with real fisheye images.  We follow previous approaches \cite{yin2018fisheyerecnet, feng2023simfir} to obtain synthetic fisheye images from the calibrated perspective images. Our synthesized fisheye images follow the distortion model introduced by Kannala \& Brandt \cite{kannala2004generic}:
\begin{equation}
    r(\theta) = k_1\theta + k_2\theta^3 + k_3\theta^5 + k_4\theta^7,
\label{eq:kb_eq}
\end{equation}
where $\theta$ denotes the angle between the ray and the optical axis, and  $\{k_i\}_{i=1}^4$ are distortion coefficients that can represent most of the real world fisheye distortion models. The change in coordinate between $(x,y)$ in the perspective image and $(x',y')$ in the fisheye image can be formulated as 
\begin{align} 
    x' &= r(\theta) \cos(\varphi), \ \ y' = r(\theta ) \sin(\varphi), \\
    \varphi &= \arctan((y - c_x)/(x - c_y)) \nonumber,
\end{align}
where $(c_x, c_y)$ is the principal point in the perspective image. We define the transformation from the perspective to the fisheye reference frame as $T$ and its inverse transformation as $T^{-1}$. Our training dataset is composed of perspective images and synthesized distorted image pairs with corresponding forward and inverse transformations.   

\noindent\textbf{Loss Function.} Inspired by AugUndo \cite{wu2024augundo} and their use of invertible transformations to preserve the supervision signal by undoing data augmentation, we propose to synthesize fisheye images from the abundance of perspective images as inputs, but undistort the output to facilitate loss computation. By remapping depth estimates of synthetic fisheye images to the perspective frame, we enable the use of high-fidelity estimates inferred by FMDEs on perspective images as supervision. Hence, we can optimize Calibration Tokens with the following self-supervised loss: 
\begin{equation}
    \label{eqn:loss_equation}
    \arg\min_{\Psi} \frac{1}{N} \sum_{n=1}^N \sum_{x \in \Omega} \log (| \tilde{d}^{(n)}(x) - T^{-1} \circ \hat{d}^{(n)}(x) | + 1),
\end{equation}
where $\tilde{d} = h_{\omega, \psi}(I)$  and $\hat{d} = h_{\omega, \psi}(T \circ I; \Phi)$ denotes the predicted depth map from the given perspective image  and synthesized fisheye image, respectively. $\Phi$ denotes the proposed trainable Calibration Token appended to the patch embeddings. 
Calibration Tokens are trained to minimize the difference between the perspective output and the fisheye output re-projected into the perspective reference frame. \cref{eqn:loss_equation} follows the log of absolute differences (logL1) proposed in \cite{peng2021excavating}, which enhances training stability and empirically outperforms L1 loss, especially in border regions where discrepancies between perspective and fisheye images are most significant (see \cref{sec:ablation} for details). 

It is important to note that attempting to instead transform the perspective depth maps outputted by FMDEs to the fisheye reference frame for the loss computation introduces information loss in the training objective. This will lead to re-projection artifacts in the supervision and result in learning inaccuracies during training. In Section A of the Supp. Mat., we further demonstrate comparison results between training in fisheye image space and perspective image space. Our training scheme is self-supervised and requires only calibrated perspective images, which can be easily obtained, making our approach both scalable and practical.


\begin{figure*} [ht]
  \centering
  \includegraphics[width=0.9\textwidth]{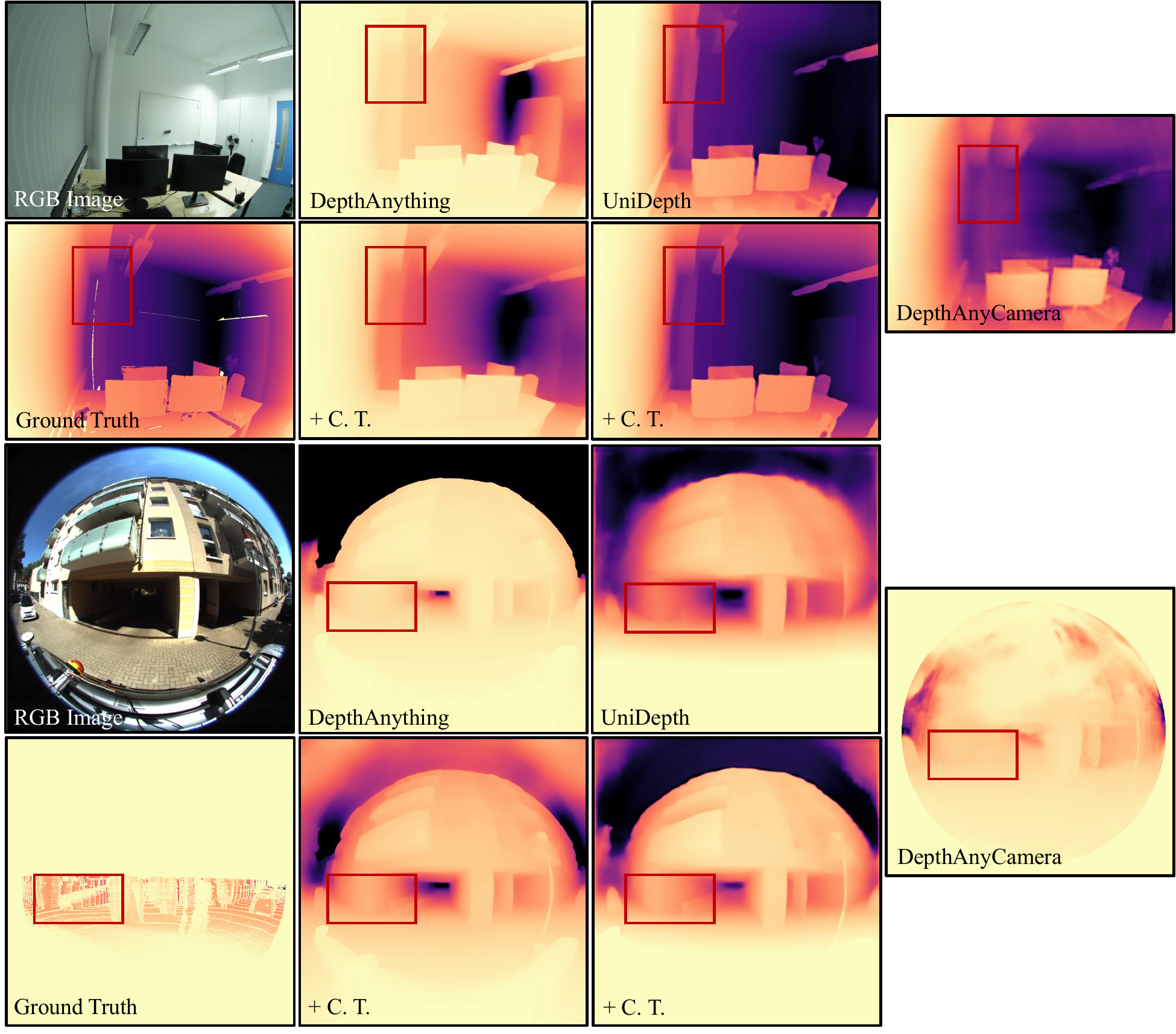}
  \vspace{-3mm}
  \caption{\textbf{Comparison on ScanNet++(Indoor) \cite{yeshwanth2023scannet++} and KITTI-360 \cite{liao2022kitti} dataset.} Qualitative comparison results on ScanNet++ and KITTI-360 datasets. Here, +C. T. indicates prediction results by appending Calibration Tokens to patch embeddings of the model located above. Calibration Tokens enable models to adapt to different fisheye cameras, especially in regions with large distortions.} \label{fig:qual_exp}
  \vspace{-3mm}
\end{figure*}

\section{Experiments}

\textbf{Datasets.} Training our Calibration Tokens requires only calibrated perspective images, enabling us to leverage significantly more data compared to training solely on fisheye images. Moreover, since our loss is computed based on comparisons with perspective image outputs, ground truth is not required for our training pipeline.

\underline{Training} datasets: \emph{NYUv2}~\cite{silberman2012indoor} has a variety of perspective indoor scenes; \emph{VOID}~\cite{wong2020unsupervised} contains indoor perspective office, classroom and stairwell scenes; \emph{IRS}~\cite{wang2021irs} contains rendered perspective scenes of home, restaurant, and store settings; \emph{Hypersim}~\cite{roberts2021hypersim} comprises photorealistic, synthetic images of indoor residential and commercial environments in a perspective reference frame. \emph{Waymo}~\cite{sun2020scalability} dataset consists of a diverse set of urban driving scenes.

\underline{Test} datasets: \emph{ScanNet++}~\cite{yeshwanth2023scannet++} offers 3D reconstructions of diverse indoor scenes, captured using laser scanning and DSLR imaging with a fisheye lens, which allows us to evaluate with real fisheye images and ground truth depth maps. \emph{KITTI-360}~\cite{liao2022kitti} includes suburban driving scenes captured with a multi-sensor setup, including fisheye cameras. Notably, it features a different field of view compared to ScanNet++~\cite{yeshwanth2023scannet++}, allowing us to assess the generalization capability of the Calibration Tokens.

\noindent \textbf{Models.} Calibration Tokens do not require specific settings and can be integrated into any model utilizing a vision transformer. We evaluate the effectiveness of our approach using by extending MiDaS \cite{ranftl2019towards}, DepthAnything \cite{yang2024depthanything}, and UniDepth \cite{piccinelli2025unidepthv2} to fisheye images. Note that we used 8 tokens per layer for each of the model experiments.

\noindent\textbf{Evaluation Metrics.} We evaluate depth prediction accuracy using standard metrics from monocular depth estimation of \emph{root mean squared error} (RMSE) and \(\delta_1\). Details on these metrics can be found in the Supp. Mat.

\noindent\textbf{Implementation Details.} We trained our Calibration Tokens based on 3 different FMDEs (MiDas\cite{ranftl2019towards}, DepthAnything\cite{ranftl2019towards}, UniDepth\cite{piccinelli2025unidepthv2}). We utilized the pretrained ViT-L backbone for MiDas \cite{ranftl2019towards} and DepthAnything \cite{ranftl2019towards}, and the ViT-S backbone for UniDepth \cite{piccinelli2025unidepthv2}. 
We trained our model on 4 NVIDIA 3090 GPUs for 40k iterations with a batch size of 16. For input, we used images in the resolution of $518\times518$. For testing, we used   $462\times616$ resolution on the ScanNet++ dataset~\cite{yeshwanth2023scannet++}, and $700\times700$ on the KITTI-360 dataset~\cite{liao2022kitti} to preserve its aspect ratio.
We also synthesize random fisheye distortions in the training images.
Our Calibration Tokens are trained with a joint dataset consisting of indoor and outdoor datasets totaling up to only 200K samples, and as shown in \cref{tab:main}, obtain comparable results on both domains with fewer samples than existing methods \cite{guo2025depth} that are trained specifically for each.

\begin{table*}[ht]
    \centering
    \caption{\textbf{Quantitative comparisons on indoors (ScanNet++) and outdoors (KITTI-360) benchmarks.} We evaluated zero-shot monocular depth estimation by incorporating trained Calibration Tokens into recent foundational monocular depth estimators models. Note: Our method uses the same training set for both the indoor and outdoor settings; whereas existing methods train separate models for each setting.}
    \label{tab:main}
    \vspace{-2mm}
    \begin{tabular}{c l l c c c}
        \toprule
        Testset & Experiment & Model & Train Dataset
        & RMSE $\downarrow$ & $\delta_1$ $\uparrow$ \\
        \midrule
        \multirow{9}{*}{{ScanNet++ \cite{yeshwanth2023scannet++}}}
        & Baseline & MiDaS \cite{ranftl2019towards} & Mix 1.4M 
         &0.506 &0.563  \\
         & \textbf{+ Calibration Tokens} & MiDaS \cite{ranftl2019towards} &  Mix 200K 
          &0.446  &0.569  \\
        & Baseline &DepthAnything \cite{yang2024depthanything} & Mix 63.5M 
         &0.731   &0.463  \\
         & \textbf{+ Calibration Tokens} &DepthAnything \cite{yang2024depthanything} & Mix 200K 
        &0.607   &0.506  \\
         &Baseline &UniDepth \cite{piccinelli2025unidepthv2} & Mix 16M 
         &0.279   &0.720  \\
         & \textbf{+ Calibration Tokens} &UniDepth \cite{piccinelli2025unidepthv2} & Mix 200K   
         &\textbf{0.244}  &\textbf{0.766}  \\
         \cmidrule(lr){2-6}
         & \multirow{3}{*}{{Comparisons}}
         & DepthAnyCamera \cite{guo2025depth}& Indoor 670K   
         &0.275   &0.761  \\ 
         & & DepthAnyCamera \cite{guo2025depth}& Mix 200K   
         &0.761   &0.255  \\
         & & FoVA-Depth \cite{lichy2024fova}& Indoor 190K 
         &{0.285} &{0.548}  \\
        \midrule
        \multirow{9}{*}{{KITTI-360 \cite{liao2022kitti}}}
        &Baseline &MiDaS \cite{ranftl2019towards} & Mix 1.4M 
          & 3.312   &0.586 \\
         & \textbf{+ Calibration Tokens} &MiDaS\cite{ranftl2019towards} & Mix 200K   
          & 2.348  &0.658  \\
         &Baseline &DepthAnything \cite{yang2024depthanything} & Mix 63.5M 
         & 2.214   &0.839  \\
         & \textbf{+ Calibration Tokens} &DepthAnything \cite{yang2024depthanything} & Mix 200K 
         & 2.043  &0.810  \\
         &Baseline &UniDepth \cite{piccinelli2025unidepthv2} & Mix 16M 
          &2.085   &0.663  \\
         & \textbf{+ Calibration Tokens} &UniDepth \cite{piccinelli2025unidepthv2} & Mix 200K  
         &\textbf{2.040} & 0.664 \\
         \cmidrule(lr){2-6}          
         & \multirow{3}{*}{{Comparisons}}
         & DepthAnyCamera  \cite{guo2025depth}& Outdoor 130K   
          &2.067   &\textbf{0.852}  \\ 
         & & DepthAnyCamera \cite{guo2025depth}& Mix 200K   
          &5.675   &0.348  \\
         & & FoVA-Depth \cite{lichy2024fova}& Outdoor 80K  
         &{3.096} &{0.632}  \\          
        \bottomrule
    \end{tabular}
    \vspace{-3mm}
\end{table*}

\subsection{Main Result}
We conduct experiments to analyze the impact of Calibration Tokens on model performance.  As a baseline, we compare our model to DepthAnyCamera \cite{guo2025depth}, the state-of-the-art monocular depth estimation (MDE) method for fisheye images. Here, we evaluate the DepthAnyCamera model using a ResNet101 backbone, and compare both its pretrained model and a model trained on our dataset for any fairness concerns. We also compare with FoVA-Depth \cite{lichy2024fova}, which is  equirectangular projection based, like DepthAnyCamera. 

\noindent\textbf{Indoor Evaluation.} Among the pretrained foundational monocular depth estimators, UniDepth achieves the best performance with our Calibration Tokens on the ScanNet++ indoor dataset as shown in \cref{tab:main}. Notably, our Calibration Tokens enable MiDAS to improve 12\% and DepthAnything to achieve a 17\% improvement in the RMSE metric compared to the model without Calibration Tokens. Similarly, UniDepth benefits from a 13\% improvement in the RMSE metric. Furthermore, compared to the comparison baselines, pretrained DepthAnyCamera and FoVA-Depth, UniDepth with Calibration Tokens surpasses their performance by 11\% and 14\% in the RMSE metric, respectively. 
DepthAnyCamera and FoVA-Depth utilize camera intrinsics for input images at test time, requiring image transformations back and forth from the equirectangular reference frame, which makes them more error-prone than our direct learning-based approach. 

\noindent\textbf{Outdoor Evaluation.} We evaluate FMDEs with our proposed Calibration Tokens against state-of-the-art methods in outdoor environments. The results show that Calibration Tokens consistently improve accuracy across different FMDEs in outdoor scenarios. Specifically, MiDaS and DepthAnything achieve improvement in the RMSE metric. UniDepth also improves 2\% in the RMSE metric, outperforming the comparison baselines. Given that the KITTI-360 dataset contains highly distorted images with a field of view exceeding 180 degrees, our Calibration Tokens demonstrate robustness across various distortion models.

Our Calibration Tokens are able to outperform DepthAnyCamera and FoVA-Depth without separate indoor and outdoor training sets, suggesting the generalization potential of our method to wide ranges of fisheye distortions. Also, the KITTI-360 ground truth points are significantly sparser and more concentrated in ground regions as compared to ScanNet++, which may explain the discrepancy in evaluation metrics. Nonetheless, our method performs comparably without needing to train specialized sets of Calibration Tokens for different fisheye models. 


\begin{figure}[t]
    \centering
    \includegraphics[width=0.8\linewidth]{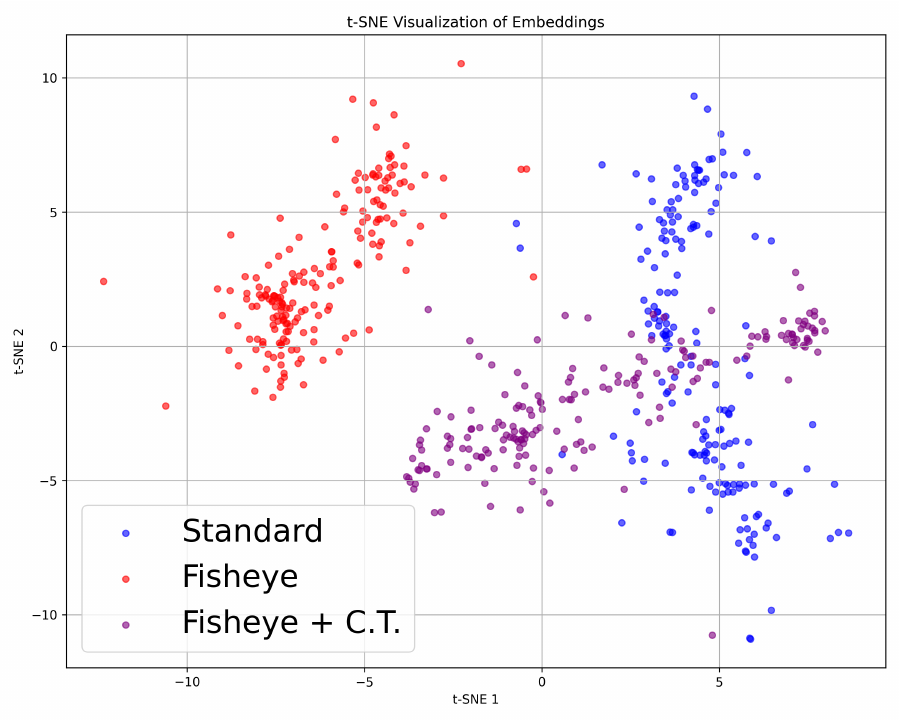}
    \vspace{-5mm}
    \caption{\textbf{t-SNE plot of fisheye and perspective embeddings.} Fisheye embeddings become closer to those of perspective images after being modulated by Calibration Tokens.} 
    \label{fig:tsne_plot}
    \vspace{-3mm}
\end{figure}

\begin{figure}[h]
    \centering
    \includegraphics[width=\linewidth]{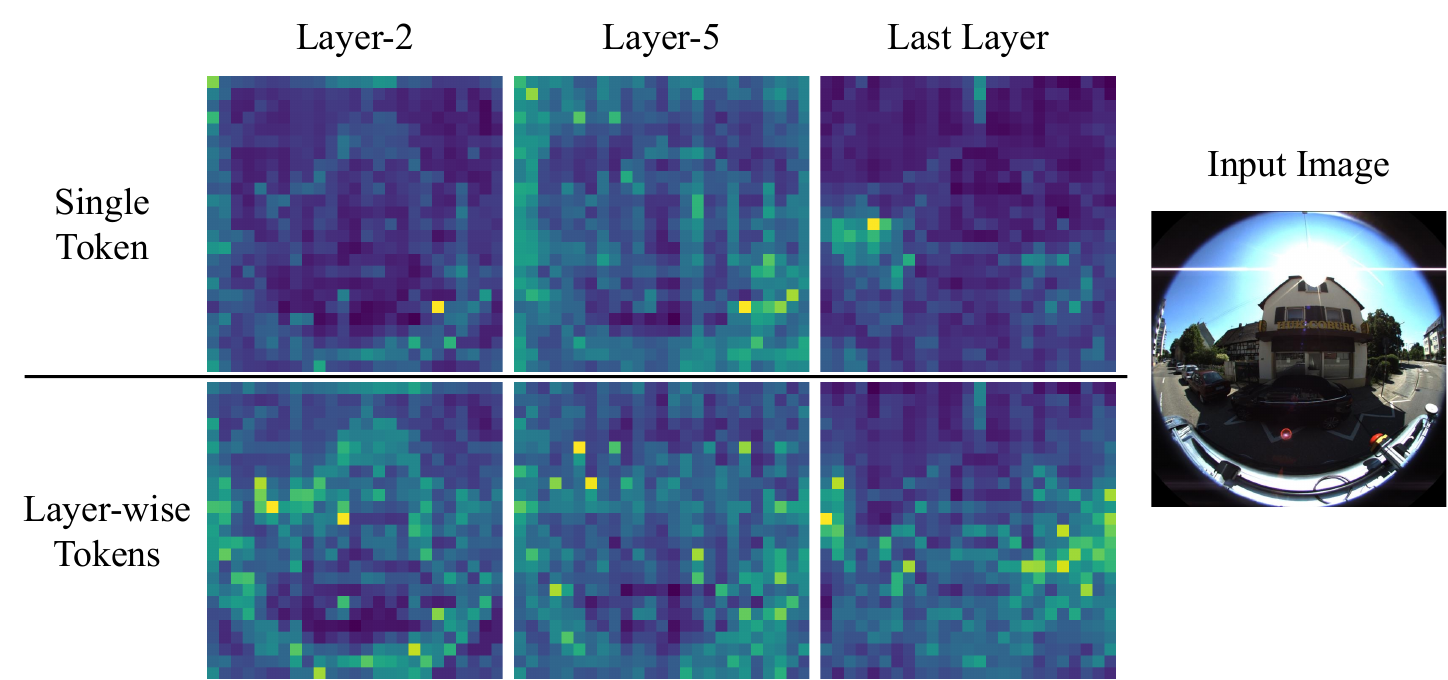}
    \vspace{-6mm}
    \caption{\textbf{Comparison of attention maps for single and multi-layer tokens.} We visualize the attention map of Calibration Tokens to the encoded patch embeddings. Calibration Tokens attend highly to distorted border regions: ``Single Token'' (top) has little effect in most layers due to lower attention as only a set of tokens are append to the input. The proposed multi-layer or ``Layer-wise Tokens'' scheme (bottom) attends to embeddings across all layers.} 
    \label{fig:token_mod}
    \vspace{-2mm}
\end{figure}

\subsection{Analysis}\label{sec:analysis}

\noindent\textbf{Feature Modulation.} To visualize how our Calibration Tokens affect fisheye embeddings,  Fig.~\ref{fig:tsne_plot} shows a two-dimensional tSNE reduction to both fisheye and perspective image embeddings from the same set of images. After adding Calibration Tokens to the fisheye embeddings, they are modulated towards the perspective image distribution.

\noindent\textbf{Layer-wise Tokens.} As observed in \cref{fig:token_mod}, when we append only a single set of tokens (``Single Token'') at the initial transformer block of the pre-trained model, the Calibration Tokens exhibit limited attention to the patch embeddings across most layers. As a result, the patch embeddings of most layers are unchanged. However, when we attach unique tokens at every layer (``Layer-wise Tokens''), we see higher attention at more layers. Thus, we opt to use the ``Layer-wise'' approach to better modulate the fisheye patch embeddings toward the distribution of perspective images. 

\begin{table}[t]
  \centering
  \caption{\textbf{Comparison results with finetuning.} We conducted experiments comparing finetuning (F.T.) with the use of Calibration Tokens (C.T.) added to the baseline model.}
  \label{tab:finetune}
  \vspace{-2mm}
  \begin{tabular}{l l c c c}
      \toprule
        Datasets &Models &Exp. &RMSE &$\delta_1$ \\
        \midrule
        \multirow{6}{*}{ScanNet++}
         & \multirow{2}{*}{MiDAS} &F.T. &2.178 & 0.129\\
         & &C.T. &0.446 & 0.569 \\
         \cmidrule(lr){2-5}
         & \multirow{2}{*}{DepthAnything} &F.T. &1.459 & 0.462\\
         & &C.T. &0.607 & 0.506 \\
         \cmidrule(lr){2-5}
         & \multirow{2}{*}{UniDepth} &F.T. &0.432 & 0.574\\
         & &C.T. &0.244 & 0.766 \\
         \midrule
        \multirow{7}{*}{KITTI-360} 
         & \multirow{2}{*}{MiDAS} &F.T. &9.289 & 0.098\\
         & &C.T. &2.348 & 0.658 \\        
         \cmidrule(lr){2-5}
         & \multirow{2}{*}{DepthAnything} &F.T. &4.362 & 0.636\\
         & &C.T. &2.043 & 0.810 \\
         \cmidrule(lr){2-5}
         & \multirow{2}{*}{UniDepth} &F.T. &3.217 & 0.403\\
         & &C.T. &2.040 & 0.664 \\         
      \bottomrule
  \end{tabular}
  \vspace{-2mm}
\end{table}

\noindent\textbf{Comparison with Finetuning.}
To further analyze the robustness of the Calibration Tokens, we conducted experiments comparing our method with a finetuning approach. We trained the model with a fixed learning rate of  \(10^{-6}\) on our synthetic fisheye dataset for the same number of iterations. As shown in \cref{tab:finetune}, the finetuning approach leads to a significant performance drop, highlighting the importance of using Calibration Tokens, which preserve the original model's training on perspective images.

\begin{table}[t]
  \centering
  \caption{\textbf{Analysis on computational cost.} We analyze the computational overhead introduced by Calibration Tokens. Values in parentheses indicate the relative increase as a percentage.}
  \vspace{-2mm}
  \label{tab:computation}
  \small
  \setlength{\tabcolsep}{9pt}
  \begin{tabular}{l@{}c c c}
      \toprule
        Models & \makecell{Model\\memory} &\makecell{Tokens\\memory} &\makecell{Inference\\time} \\
        \midrule 
        MiDAS &1.7G & 0.8M(0.05\%) & 0.6ms(0.8\%) \\
        DepthAnything &1.7G & 0.8M(0.05\%) &0.8ms(0.8\%) \\
        UniDepth &0.7G & 0.2M(0.02\%) & 0.4ms(0.7\%) \\

      \bottomrule
  \end{tabular}
  \vspace{-3mm}
\end{table}

\noindent\textbf{Computational Cost.}
\cref{tab:computation} shows the impact of Calibration Tokens on computational costs across different FMDEs. Incorporating Calibration Tokens results in only a 0.05\% and 0.02\% increase in memory usage, less than 1 MB and a 0.8\% and 0.7\% increase in inference time, with an added latency of less than 1 ms. This analysis highlights the efficiency of our proposed Calibration Tokens.

\subsection{Ablation Study}\label{sec:ablation}
We conducted an ablation study on different Calibration Token configurations to validate our contributions. Note that ``Single Token'' refers to a single set of Calibration Tokens appended in the first layer of the vision transformer without removal, with L1 loss applied. In this configuration, $\phi \in \mathbb{R}^{ M \times F}$ as opposed to $\Phi \in \mathbb{R}^{L \times M \times F}$ in the layer-wise setting. The ablation study on the ScanNet++ and KITTI-360 datasets is performed using the UniDepth model.


\begin{table}[t]
  \centering
  \caption{\textbf{Ablation study.} We ablate the training objective and modulate scheme for our proposed Calibration Tokens.}
  \vspace{-2mm}
  \label{tab:ablation}
  \begin{tabular}{l l c c}
      \toprule
        Dataset &Method &RMSE &$\delta_1$ \\
        \midrule
        \multirow{3}{*}{ScanNet++}
        & Single token &0.260 & 0.741\\
         & + LogL1 Loss &0.254 & 0.752 \\
         & + Layer-wise Tokens & \textbf{0.244}  &\textbf{0.766}\\
         \midrule
        \multirow{3}{*}{KITTI-360} 
        & Single token &2.085 & 0.656\\
         & + LogL1 Loss &2.065  & \textbf{0.665}\\
         & + Layer-wise Tokens &\textbf{2.040} & 0.664\\
      \bottomrule
  \end{tabular}
  \vspace{-3mm}
\end{table}

\noindent\textbf{LogL1 Loss.} 
We observed stable improvements with LogL1 loss compared to baseline L1 loss. As shown in \cref{tab:ablation}, the LogL1 loss improves both metrics across indoor and outdoor datasets. Qualitative comparisons between L1 and LogL1 objectives are shown in the Supp. Mat.

\noindent\textbf{Layer-wise Tokens.} \cref{tab:ablation} demonstrates the advantages of using layer-wise tokens over a single set of Calibration Tokens in the first layer. Even when the same number of tokens is fed to the visual transformer layers, we observed a significant improvement in the contribution of layer-wise tokens. This supports our hypothesis about how the influence of Calibration Tokens diminishes through a forward pass as observed in our experiments by appending a single set of tokens at the first layer. \cref{fig:token_mod} visualizes attention.

\section{Discussion}
Calibration Tokens enable FMEs to adapt to images captured by fisheye cameras. Empirically, our method improves on monocular depth estimation on fisheye cameras.
While our method trains only one set of tokens for both indoor and outdoor settings, our promising results motivates this as a general approach to adapting vision foundational models. 
Furthermore, a convenience afforded by our method is in the reuse and backward-compatibility of FMDEs with perspective images. This reduces the operational overhead of multi-camera systems by enabling a single FMDE to handle multiple camera inputs -- adding cameras become as easy as appending tokens.

\noindent\textbf{Limitations.} While we offer a light-weight method of extending FMDEs to fisheye images, its success inherently depends on the quality and representational power of the underlying FMDEs. If the pretrained model struggles with certain 3D scenes or lighting conditions for perspective images, then these issues carry over. Nonetheless, as novel FMDEs emerge, our framework can be readily transferred to new models using transformer-based architectures.



\paragraph{Acknowledgments}
This work is supported by NSF 2112562 Athena AI Institute, the Rosenfeld Science Scholars Program, and gift funding from Google.

{
    \small
    \bibliographystyle{ieeenat_fullname}
    \bibliography{main,monocular_depth,visionlab, sup}
}

\newpage

\appendix

\twocolumn[{
 \centering
  {\Large{\textbf{ 
  Extending Foundational Monocular Depth Estimators to Fisheye Cameras \\ with Calibration Tokens \vspace{0.5cm}
  \\ SUPPLEMENTARY MATERIAL}}}
  \vspace{1cm}
}]

\begin{table}[t]
    \caption{\textbf{Additional experiments.} }
    \vspace{-0.2cm}
    \centering
    \label{tab:supp_mat_table}
    \setlength{\tabcolsep}{2pt}
    \begin{tabular}{c l l c c}
        \midrule
        \hspace{20pt} & {Experiment}
        & {Model} & {RMSE$\downarrow$} 
        & {$\delta_1$$\uparrow$} \\
        \midrule
        \multirow{4}{*}{\adjustbox{angle=90,valign=m}{ScanNet++}}
        & Self-supervised (ours)
        & {UniDepth}   
        & \underline{0.244} &\underline{0.766}  \\   
        & Supervised (ours) 
        & UniDepth
        & \textbf{0.242} &\textbf{0.769}  \\
        \cline{2-5}      
        \addlinespace[0.5em]
        & Fisheye space
        & UniDepth
        & 0.280 &0.755  \\        
        & Same token added
        & UniDepth        
        & 0.290 &0.752  \\          
        \midrule
        \multirow{4}{*}{{\adjustbox{angle=90,valign=m}{KITTI-360}}}
        & {Self-supervised (ours)}
        & {UniDepth}   
        &{\underline{2.040}} &{\textbf{0.664}}  \\     
        & Supervised (ours)
        & {UniDepth}
        &{\textbf{1.994}} &{\underline{0.651}}  \\  
        \cline{2-5}
        \addlinespace[0.5em]
        & Fisheye space
        & {UniDepth}
        &{2.110} &{0.618}  \\       
        & Same token added
        & {UniDepth }
        &{2.062} &{0.631}  \\       
        \midrule
    \end{tabular}
\end{table}

\section{Additional Experiments}

To further validate our claims and design choices, we evaluated the performance of some other possible designs, which can be seen in Tab.~\ref{tab:supp_mat_table}.

\begin{figure} [h]
    \centering
    \includegraphics[width=0.7\linewidth]{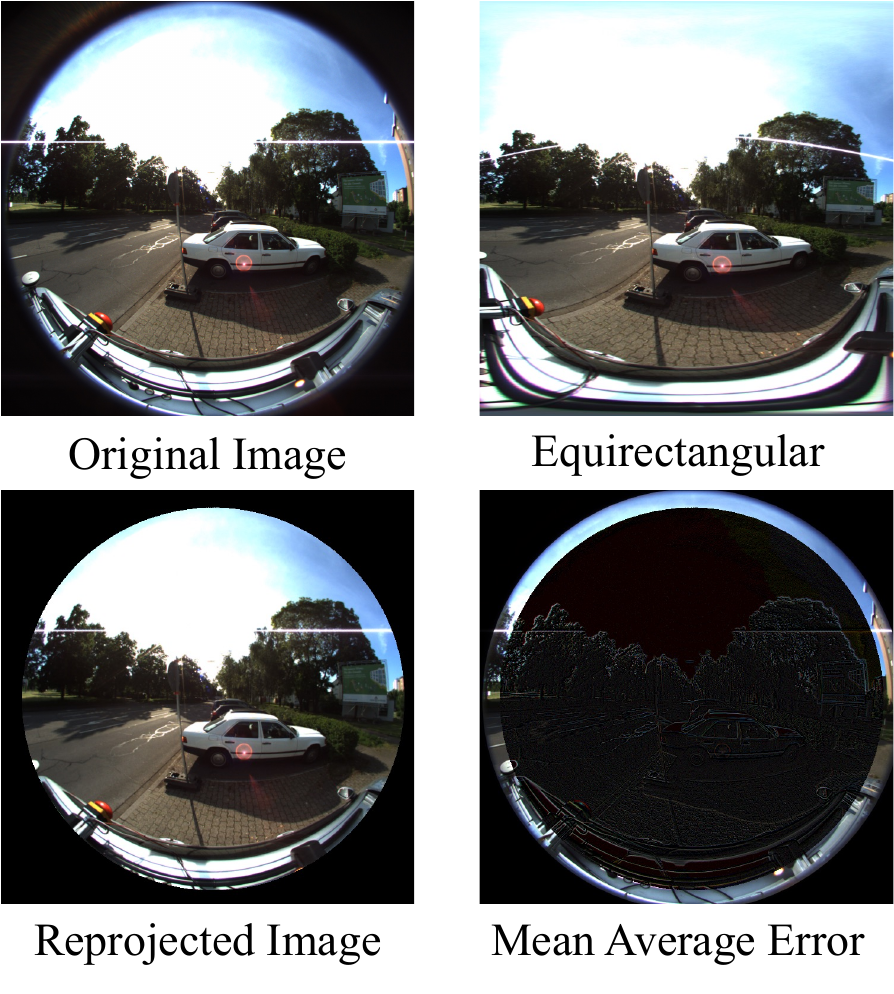}
    \caption{\textbf{Visualization of lossy training objective.}}    \label{fig:transformation_artifacts}
\end{figure}

\noindent\textbf{Fisheye Frame Loss.} In the main paper, we claimed that computing loss in the fisheye reference frame would perform worse because we would need to transform the perspective output, which would give us a lossy training objective. We have validated that claim with another experiment in the table. Furthermore, Fig.~\ref{fig:transformation_artifacts} shows the information loss caused by distorting to the equirectuangular space, which is used by some baseline methods. In this example with an image from KITTI-360, there is a 17.23\% loss in the image pixels.

\noindent\textbf{Same Token Added.} In addition to the "Layer-wise" and "Single Token" approaches for adding our calibration tokens that we discussed in the main paper, we tried taking the same token, but adding and removing it after each transformer block, so it remains unchanged for each transformer block. We found that this approach still does not outperform the "Layer-wise" approach. 

\noindent\textbf{Supervised Loss.} Because our loss is self-supervised (using output from a pretrained model as the training objective), we also evaluate the performance of our method when training with perspective ground truth instead of the perspective model output. As expected, there is a slight performance increase. However, it would be more cost-effective to use the self-supervised approach because the improvement is limited, especially in the indoor setting. This further validates the robustness of the baseline foundation model for perspective images.

\begin{figure}[h]
    \centering
    \includegraphics[width=\linewidth]{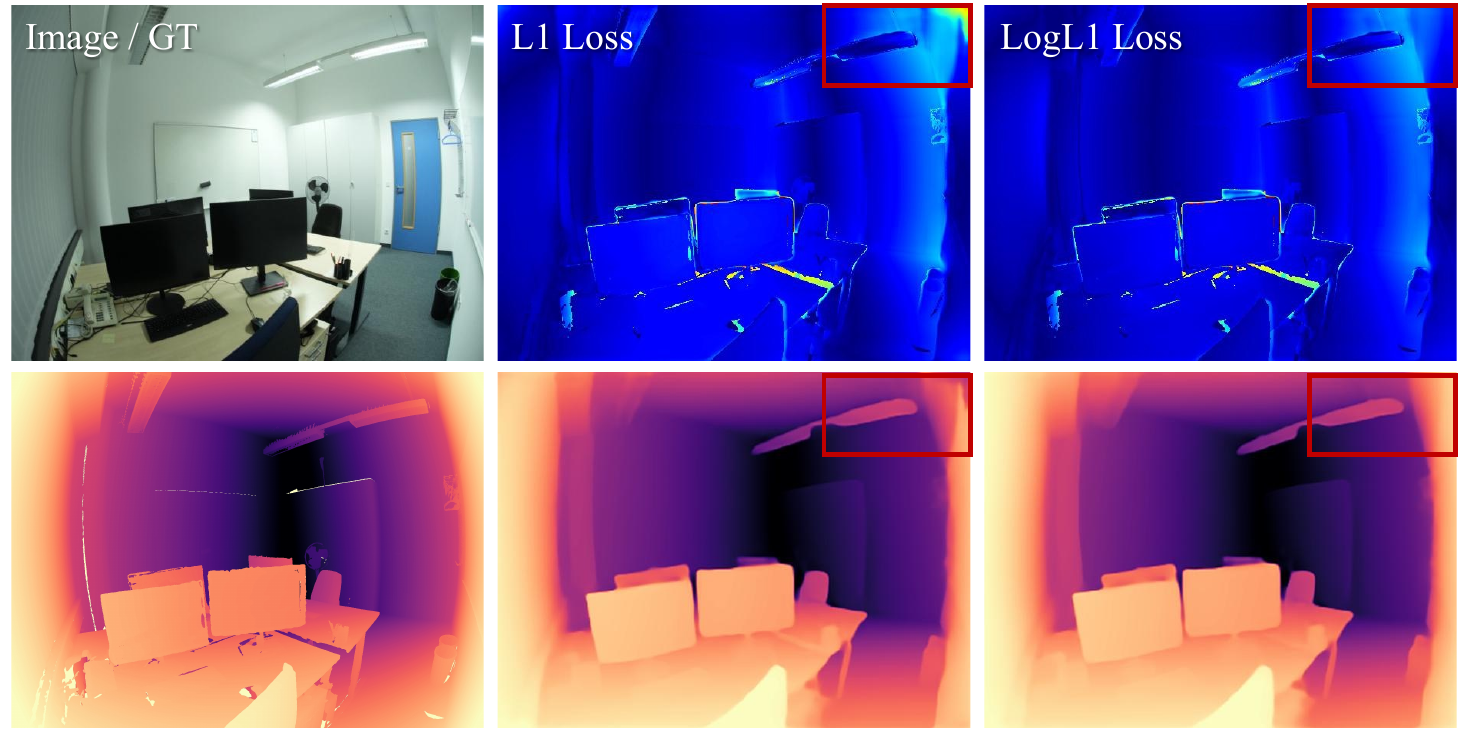}
    \caption{\textbf{Validation on LogL1 loss.} We evaluate the effectiveness of our LogL1 loss by comparing a single-layer token baseline with an additional LogL1 loss. Incorporating LogL1 loss helps model to mitigate artifacts in the highlighted border regions of fisheye images, leading to improved visual consistency.} 
    \label{fig:ablation}
\end{figure}

\begin{figure*}[!h]
  \centering
  \includegraphics[width=\textwidth]{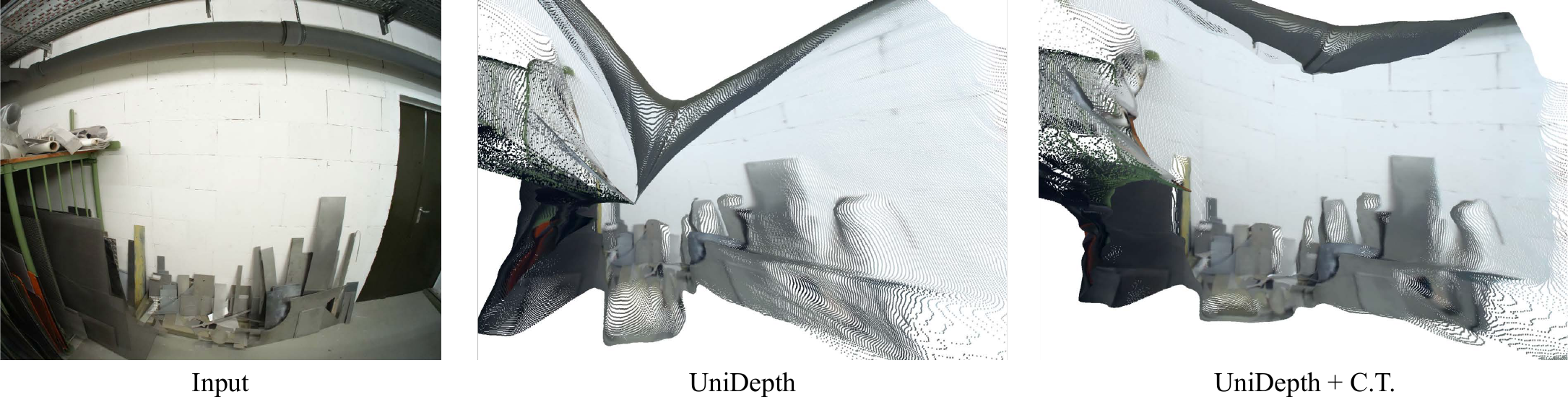}
  \caption{\textbf{3D reconstruction result of UniDepth predictions on ScanNet++ dataset.}}
  \label{fig:point_clouds}
\end{figure*}

\begin{figure*}[h]
  \centering
  \includegraphics[width=\textwidth]{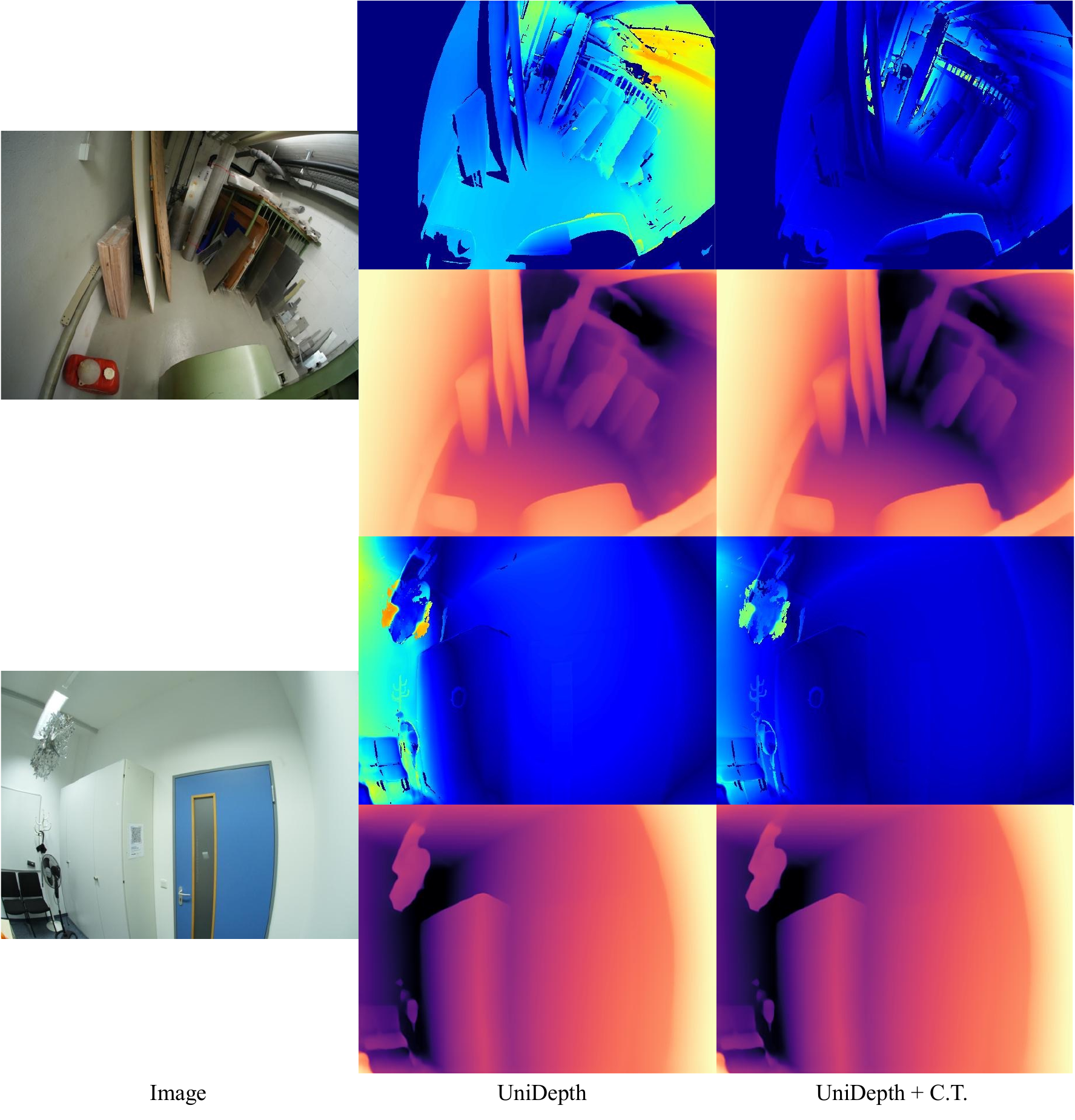}
  \caption{\textbf{Additional comparison results on ScanNet++ dataset.}}
  \label{fig:additional_exp1}
\end{figure*}

\begin{figure*}[h]
  \centering
  \includegraphics[width=\textwidth]{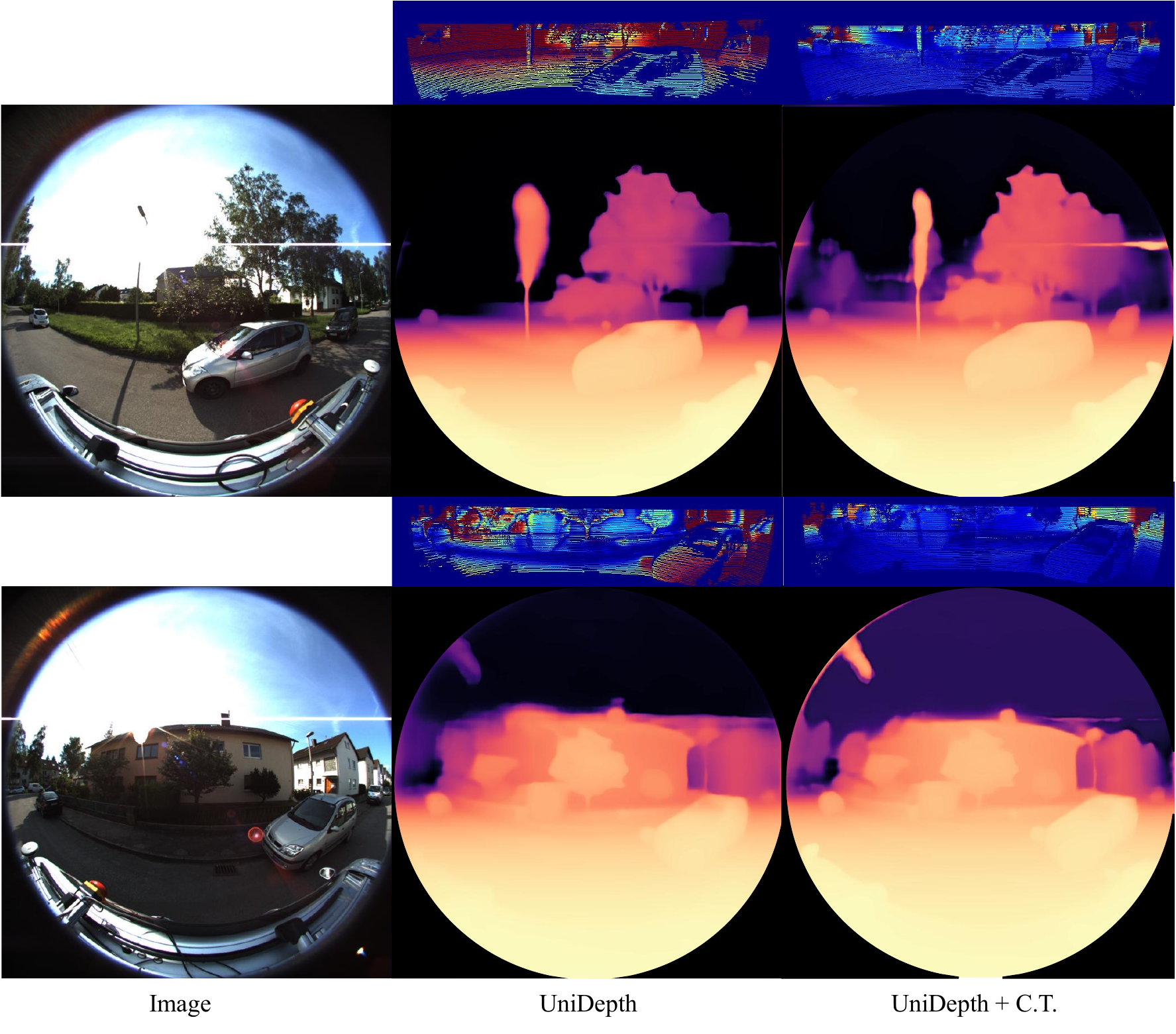}
  \caption{\textbf{Additional comparison results on KITTI-360 dataset.}}
  \label{fig:additional_exp2}
\end{figure*}

\noindent\textbf{Additional Qualitative Results.} We further demonstrate our contribution with the 3D reconstruction results as shown in \cref{fig:point_clouds}. This result provides evidence of our contribution toward foundational model latent embeddings to be aligned to fisheye images with our fully self-supervised training. Additionally, we provide qualitative results to validate our LogL1 loss. As can be seen with the \cref{fig:ablation}, the logL1 loss helps the model mitigate the impact of artifacts caused by severe distortions, leading to more stable improvements on fisheye images, as reflected in the depth map and error map results. \cref{fig:additional_exp1} and \cref{fig:additional_exp2} visualize the depth estimation comparison with and without the calibration token (C.T.) on the ScanNet++ and KITTI-360 datasets, respectively.

\begin{table}[h]
\centering
\resizebox{\columnwidth}{!}{%
\begin{tabular}{l l}
\midrule
\textbf{Metric} & \textbf{Definition} \\
\midrule
RMSE\,$\downarrow$ &
\(\displaystyle 
\sqrt{\frac{1}{|\Omega|} \sum_{p \in \Omega} 
\bigl(\hat{d}(p) - d(p)\bigr)^2}
\)
\\[0.6em]
\(\delta_1\)\,$\uparrow$ &
\(\displaystyle 
\frac{1}{|\Omega|} \sum_{p \in \Omega} 
\mathbf{1}\!\Bigl(
\max\!\Bigl(\frac{\hat{d}(p)}{d(p)}, 
\, \frac{d(p)}{\hat{d}(p)}\Bigr) 
< 1.25^1
\Bigr)
\)
\\
\midrule
\end{tabular}%
}
    \caption{
        \textbf{Error metrics for depth estimation.} These evaluation metrics compute the error between predicted depth values $\hat{d}(x)$ and ground truth depth values $d(x)$.}
\label{tab:depth_metrics}
\end{table}

\section{Additional Details}
\subsection{Foundational Depth Estimation Models}

\noindent\textbf{MiDAS, DepthAnything-V1(ViT-L).}
Following the pipeline of \cite{ranftl2019towards,yang2024depthanything}, these models utilizes a Vision Transformer Large encoder and a specialized decoder head for single-view depth estimation. Its training covers a massive corpus of perspective images drawn from both indoor and outdoor domains, aiming at robust zero-shot performance. Despite strong generalization within pinhole-camera distributions, it lacks dedicated mechanisms for counteracting severe lens distortions (e.g., fisheye or panoramic).

\noindent\textbf{UniDepth-V2(ViT-S).}
UniDepth-V2 \cite{piccinelli2025unidepthv2} leverages a Vision Transformer Small backbone, paired with a camera self-prompting routine to address moderate discrepancies in intrinsic parameters. However, when confronted with extreme distortions typical of ultra-wide or fisheye lenses, it is insufficient to recover geometry reliably. In both cases, we demonstrate how a small set of learnable calibration tokens (see main paper) can bridge the gap from perspective to fisheye images without retraining the full models.

\subsection{Datasets}
We provide further details on the datasets used for both training and testing.

\emph{Training Datasets:}
\textbf{NYU Depth V2}~\cite{silberman2012indoor} (“NYUv2”) consists of 464 diverse indoor scenes (e.g., living rooms, offices). It contains about 400{,}000 aligned RGB--depth pairs at 640\(\times\)480 resolution. Following standard practice, approximately 1{,}500 depth points are chosen in each map via the Harris corner detector~\cite{harris1988combined}. NYUv2 is a common benchmark for indoor depth tasks and serves here as one of our primary training sets.

\noindent\textbf{IRS}~\cite{wang2021irs} compiles a large number of synthetic indoor environments, from small apartments to commercial interiors—each scene offering ground-truth depth rendered at resolutions comparable to 640\(\times\)480. Its scale (up to 103{,}316 frames) and variety of virtual layouts supplement real data.

\noindent\textbf{VOID}~\cite{wong2020unsupervised} (Visual Odometry with Inertial and Depth) features about 58{,}000 frames taken in hallways, classrooms, and shared spaces, each accompanied by a sparse depth map at roughly 0.5\% density (\(\approx\)1{,}500 points).

\noindent\textbf{Hypersim}~\cite{roberts2021hypersim} is a photo-realistic synthetic dataset offering about 77{,}400 RGB--depth pairs. These scenes incorporate meticulously rendered geometry and lighting across various architectural styles (e.g., residential, museum-like structures). Hypersim’s controlled yet visually realistic design helps our model see a wide spectrum of interior layouts even before encountering real-world test sets.

\noindent\textbf{Waymo Open Dataset}~\cite{sun2020scalability} contributes \(\sim\)230{,}000 camera--LiDAR frames across urban and suburban roads. Though heavily used for self-driving applications (e.g., detection, tracking), we leverage it here to extend our token training beyond the pure indoor scenario. The inclusion of Waymo frames exposes our method to outdoor scenes with larger view ranges and more complex lighting.

\emph{Testing Datasets:} 
Our proposed approach is primarily evaluated on two real-world datasets that each incorporate fisheye or wide-FOV imaging. \noindent\textbf{ScanNet++}~\cite{yeshwanth2023scannet++} is an extended collection of indoor RGB-D sequences, building on the popular ScanNet dataset but augmented with additional scenes and fisheye captures. We use the fisheye depth estimation ground truth to verify how our framework handles substantial lens distortion indoors. 

\noindent\textbf{KITTI-360}~\cite{liao2022kitti} is an outdoor dataset focusing on large-scale mapping and autonomous driving. It contains 360$^\circ$ fisheye cameras and high-grade LiDAR depth. Scenes encompass suburban roads, semi-rural stretches, and detailed 3D annotations. Testing on KITTI-360 lets us measure the ability of our approach to generalize to wide-FOV imagery in challenging real-world driving contexts.

\subsection{Implementations}
All experiments used the same training hyperparameters: Adam optimizer with learning rate of \(10^{-4} \) and \(\beta_1 = 0.9, \beta_2=0.999\).
For random fisheye distortion synthesis, we leveraged the polynomial distortion model introduced by Kannala \& Brandt ~\cite{kannala2004generic}, using four distortion parameters (i.e., $N_k=4$) within the range of $[-1.0, -0.01]$.

\subsection{Evaluation Metrics}
For the evaluation, we used metrics proposed by Eigen et al.\cite{eigen2014depth}. Since our focus is on adapting monocular depth estimation to different visual modalities, we measure relative depth estimation performance to mitigate the gap introduced by fisheye images. This is crucial, as foundation models often suffer from a loss of general performance in such cases.
\cref{tab:depth_metrics} provides detailed equations used for evaluation. The \emph{root mean squared error} (RMSE) measures deviation in the linear depth space. We further report a threshold-based accuracy, \(\delta_1\), which represents the percentage of pixels whose predicted depth is within a tight bound of the ground-truth depth.

\section{Discussion}

Spatial applications are typically deployed on platforms (e.g., robots, autonomous vehicles, extended reality headsets) with multi-camera systems. Naturally, data collection is done on a specific platform that may differ from those used during deployment. This introduces a domain or covariate shift between the training and testing distributions. The focus of this paper is on the covariate shift introduced by fisheye cameras, which are common to many spatial platforms. While we demonstrate our method on monocular depth estimation \cite{godard2017unsupervised,lao2024depth,lao2024sub,upadhyay2023enhancing, wang2025ode,wong2019bilateral,wong2020targeted,wu2024augundo,fei2019geo,zeng2024priordiffusion,zeng2024rsa,zeng2024wordepth,zhou2017unsupervised}, it is just one of many perception tasks that are affected by this covariate shift: We see further applications in optical flow \cite{lao2017minimum,lao2018extending,lao2019minimum,lao2024diffeomorphic,sun2018pwc,teed2020raft,zhang2024adaptive,zhang2024heteroscedastic}, semantic segmentation \cite{chen2017deeplab,wong2015one,wong2017exploiting,wong2021small,han2023spatial, xie2021segformer,zhang2025progressive}, image restoration \cite{ba2022not,zamir2022restormer,zhang2023weatherstream} and stereo \cite{berger2022stereoscopic,gu2020cascade,wang2021patchmatchnet,wong2021stereopagnosia,xu2020aanet}. Further, many perception tasks follow the convention of projecting different sensor modalities onto the image reference frame for fusion. We envision our method to be applicable towards perception tasks on multi-sensor platforms, including 3D objection  \cite{xia2023quadric,xie2023sparsefusion} with camera and LiDAR and 3D reconstruction with camera and LiDAR \cite{chancan20253d,ezhov2024all,liu2022monitored,park2020non,park2024test,chung2025eta, wong2020unsupervised,wong2021adaptive,wong2021learning,wong2021unsupervised,yang2019dense} or radar \cite{rim2025radar,singh2023depth}. Finally, we see a connection between our method and continual learning \cite{chen2024uncle,mccloskey1989catastrophic,rim2025protodepth,thrun1995learning,yang2024binding} as our method aims extend to models to different cameras, e.g. perspective to fisheye, instead of 3D scenes while maintaining previously learned information, e.g., backward-compatibility.

\break

\newpage

\clearpage



\end{document}